\documentclass[lettersize,journal]{IEEEtran}
\usepackage{amsmath,amsfonts}
\usepackage{algorithmic}
\usepackage{array}
\usepackage[caption=false,font=normalsize,labelfont=sf,textfont=sf]{subfig}
\usepackage{textcomp}
\usepackage{hyperref}
\usepackage{booktabs}
\usepackage{cite}
\usepackage{stfloats}
\usepackage{makecell}
\usepackage{multirow}
\usepackage[most]{tcolorbox}
\usepackage{url}
\usepackage{verbatim}
\usepackage{graphicx}
\def\BibTeX{{\rm B\kern-.05em{\sc i\kern-.025em b}\kern-.08em
    T\kern-.1667em\lower.7ex\hbox{E}\kern-.125emX}}
\usepackage{balance}

\definecolor{darkblue}{rgb}{0, 0, 0.5}
\hypersetup{colorlinks=true,citecolor=darkblue, linkcolor=darkblue, urlcolor=darkblue}
\definecolor{mygreen}{RGB}{46,139,87} 

\begin{document}
\title{Understanding Before Reasoning: Enhancing Chain-of-Thought with Iterative Summarization Pre-Prompting}
\author{Dong-Hai Zhu, Yu-Jie Xiong, Jia-Chen Zhang, Yan Xu, Xi-Jiong Xie, Chun-Ming Xia
	\thanks{
		Dong-Hai Zhu, Yu-Jie Xiong, Jia-Chen Zhang, Yan Xu, and Chun-Ming Xia are with the School of Electronic and Electrical Engineering, Shanghai University of Engineering Science, Shanghai, China. (Corresponding author:Yu-Jie Xiong.)
		
		Xi-Jiong Xie is with the School of Information Science and Engineering, Ningbo University, Ningbo, China.}
}
\markboth{IEEE Transactions on Audio, Speech and Language Processing, Vol. 1, No. 1, January 2021}
{Dong-Hai Zhu, \MakeLowercase{\textit{(et al.)}: Understanding Before Reasoning: Enhancing Chain-of-Thought with Iterative Summarization Pre-Prompting}}
\maketitle

\begin{abstract}
Chain-of-Thought (CoT) is the dominant paradigm applied in Large Language Models (LLMs) to enhance their capacity for complex reasoning.
It guides LLMs to demonstrate the problem-solving process through a chain of reasoning steps, rather than requiring LLMs to generate the final answer directly.
Despite its success, CoT encounters difficulties when key information required for the reasoning process is either implicit or missing.
It primarily stems from the fact that CoT emphasizes the stages of reasoning, while neglecting the critical task of gathering and extracting essential core information in the early stage.
In this paper, we propose a pre-prompting methodology called Iterative Summarization Pre-Prompting ($\text{ISP}^{2}$), which can effectively refine the reasoning ability of LLMs when key information is not explicitly presented. 
First, entities and their corresponding descriptions are extracted to form potential key information pairs from the question.
Next, we introduce the reliability rating to assess the reliability of these information pairs, prioritizing those with lower rankings which retain effectiveness for problem-solving but require further knowledge extraction. 
Then, the two lowest-ranked pairs, which demonstrate problem-solving capacity yet underdeveloped knowledge associations, are merged through summarization.
The summarization process synthesizes two pairs into a consolidated description, exposing implicit relationships that exist between information pairs.
The reliability rating and summarization are iteratively applied to guide the generation of a unique information pair.
Finally, the obtained unique information pair, along with the original question, is fed into LLMs for reasoning, resulting in the final answer.
Extensive experiments are conducted to validate the effectiveness of the proposed method. 
The results show that, compared to existing methods, our approach yields a 7.1\% improvement in performance.
In summary, unlike traditional prompting methods, $\text{ISP}^{2}$ adopts an inductive approach with pre-prompting.
It demonstrates good plug-and-play performance and can theoretically be applied to improve performance across all reasoning frameworks.
The code is available at: \url{https://github.com/X-Lab-CN/ISP}.
\end{abstract}


\section{Introduction}
\IEEEPARstart{L}{arge} Language Models (LLMs) have made significant strides in Natural Language Processing (NLP) tasks, such as question answering, automatic summarization, and machine translation \cite{openai2023gpt4, chowdhery2022palm, touvron2023llama, touvron2023llama2, huang2022large, zhao2023survey}.
However, merely scaling up model parameters has shown limited effectiveness in bridging the gap between LLM reasoning capabilities and human-level performance.
Chain-of-Thought (CoT) has emerged as a promising approach to enhance reasoning processes.
By decomposing complex problems into intermediate steps through prompts like "let's think step by step" \cite{Kojima2022LargeLM} or demonstration-based learning \cite{wei2022chain}, CoT has demonstrated measurable improvements in zero-shot and few-shot reasoning benchmarks.
Nevertheless, the effectiveness of CoT methods depends on the complexity of the problem.
While CoT provides more interpretable prediction paths for moderately complex tasks, it faces limitations when handling highly complex reasoning scenarios.
Challenges such as the frequent overlooking of critical information can lead to unclear guiding strategies, ultimately impacting the quality of the answers.

Simon et al.\cite{Simon1978InformationProcessingTO}’s theory provides valuable insights to better understand and solve problems.
The theory establishes a formal framework for problem-solving analysis, where human cognition constructs problem spaces through systematic operator applications and goal formalization.
With an understanding of the current context, humans use heuristic strategies to summarize and refine their thoughts, gradually approaching the solution to the goal.
Building upon Simon's foundational work in information processing theory, this paper applies these concepts to the reasoning of LLMs and introduces Iterative Summarization Pre-Prompting ($\text{ISP}^{2}$).
$\text{ISP}^{2}$ operationalizes three core processes aligned with classical problem-solving theory: (1) formalization of state transition operators via task-driven input filtering, (2) rule-based evaluation of relational mappings under dynamic constraints, and (3) goal-bound hierarchical consolidation with termination conditions. 
The structured method addresses the limitations of implicit knowledge assumptions in LLMs by explicitly constructing problem representations through observable operations, thereby maintaining theoretical consistency with Simon's information processing work.

Specifically, $\text{ISP}^{2}$ is a pre-prompting method applied before CoT, allowing the LLMs to summarize more comprehensive knowledge to assist in reasoning. 
In Figure \ref{fig8}, we illustrate the differences between two paradigms for enhancing CoT reasoning: one is the mainstream approach that augments CoT during the reasoning stream, and the other is our method of enhancing CoT through pre-prompting.
$\text{ISP}^{2}$ enhances CoT through pre-prompting by systematically enriching the input context before reasoning. 
It coordinates three key LLM steps: adaptive extraction of candidate information, reliability rating of information pairs, and iterative summarization for knowledge understanding. 
These steps include summarizing and integrating relevant information, as well as formulating strategies before tackling intricate real-world reasoning tasks. 
By engaging in these pre-prompting steps, LLMs can better explore and understand the nuances of complex problems, thereby improving their ability to perform sophisticated reasoning.
Testing with GPT-3.5 Turbo shows that inserting $\text{ISP}^{2}$ leads to a significant performance improvement, with increases of 7.1\% and 8.1\%, respectively.
The average performance score of $\text{ISP}^{2}$ with CoT reaches 79.43, surpassing other SOTA methods with plug-and-play capabilities.
The results validate the effectiveness of integrating formal problem space principles into modern LLM prompting strategies.
In these processes and results, our main contributions are as follows:

\begin{figure}
	\centering
	\includegraphics[width=3.5in]{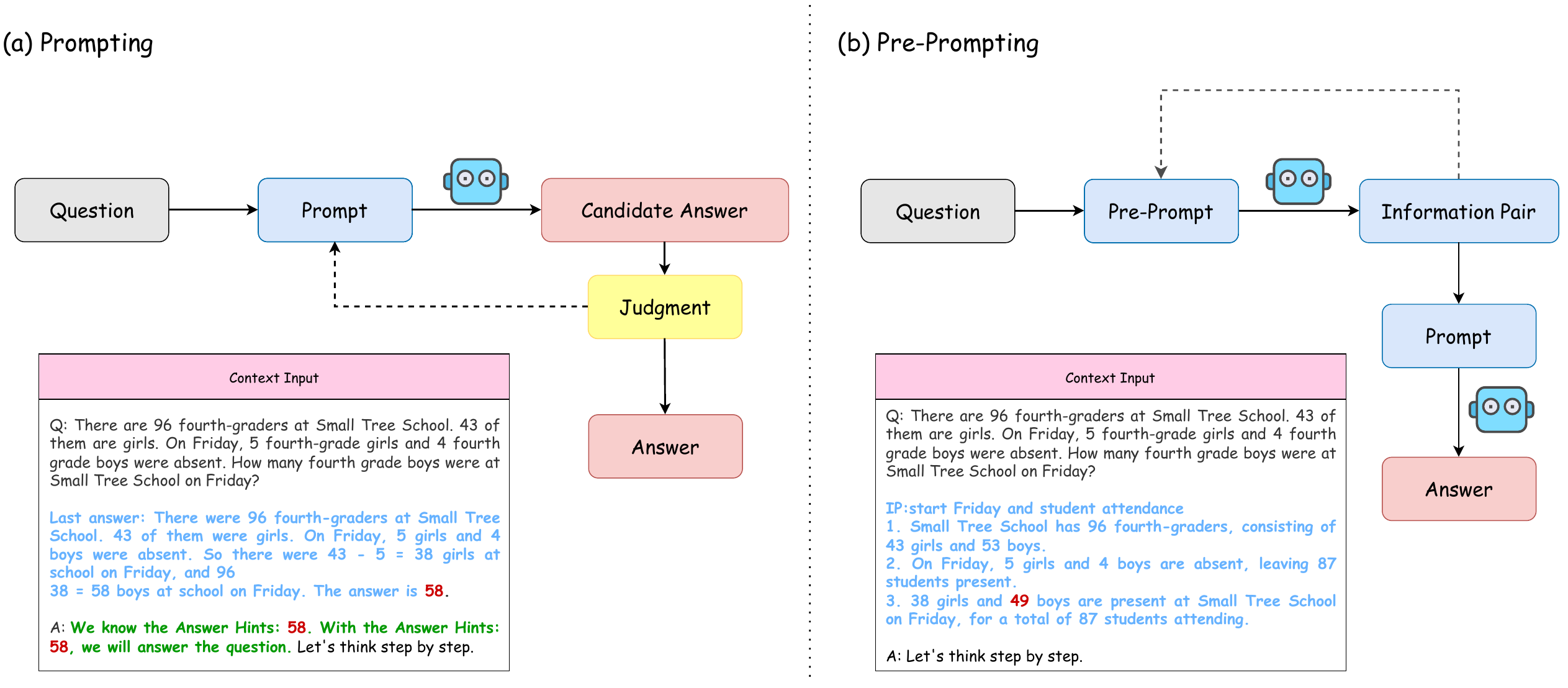}
	\caption{The difference between mainstream prompt enhanced CoT and pre-prompt enhanced CoT. The left part illustrates the mainstream approach, which augments CoT during the reasoning process by dynamically guiding the model's inference steps. The right part demonstrates the pre-prompt method, which enhances CoT through pre-prompting by systematically enriching the input context before reasoning. The distinction highlights how pre-prompting shifts the focus from in-process guidance to upfront contextual refinement, enabling more robust reasoning from the outset.}
	\label{fig8}
\end{figure}

\begin{itemize}
	\item $\text{ISP}^{2}$ is a pre-prompting method that can be seamlessly integrated into various CoT methods to enhance their reasoning performance. Furthermore, it achieves top performance among SOTA plug-and-play methods.
	\item We extend information extraction to iterative generation and reliability rating, creating a new process for step-by-step information integration in complex scenarios.
	\item The results demonstrate that $\text{ISP}^{2}$ achieves outstanding performance across various LLM environments, from closed-source models (GPT-3.5 Turbo, GPT-4o-mini) to open-source models (LLaMA2/3, DeepSeek-Distill), verifying its compatibility with diverse parameter regimes and transformer-based architectures.
\end{itemize}

\section{Related Work}

\subsection{Chain-of-Thoughts Prompting}
Wei et al.\cite{wei2022chain} emphasize the importance of deriving conclusive answers through multi-step logical pathways by introducing the concept of Chain-of-Thoughts (CoT) reasoning. 
The method demonstrates that reasoning abilities can be elicited through a series of thoughtful steps.
Kojima et al.\cite{Kojima2022LargeLM} discover that simply adding the phrase "let's think step by step" in prompts allows LLMs to perform zero-shot logical reasoning without any additional human prompts.
Subsequently, Wang et al.\cite{wang2023selfconsistency} introduce Self Consistency (SC) to replace the greedy decoding strategy. 
Zhang et al.\cite{zhang2023automatic} construct an automatic CoT framework based on the problem, eliminating the instability of manual prompts. 
Fu et al.\cite{fu2023complexitybased} employ complexity-based multi-step reasoning estimation to execute CoT.
Yao et al.\cite{NEURIPS2023_271db992} propose Tree-of-Thoughts (ToT), which introduces deliberation into decision-making by considering multiple reasoning paths.
Xu et al.\cite{xu2024rereading} enhance the model's understanding by re-reading the question. 
These studies underscore the importance of CoT in enhancing the reasoning and planning capabilities of LLMs in complex scenarios.
Despite, CoT still requires further refinement in complex scenarios involving more complex problems.

\subsection{In-Context Learning}
In-context learning (ICL) enables LLMs to make predictions based on input examples without updating model parameters.
Brown et al.\cite{10.5555/3495724.3495883} introduce this concept in GPT-3, demonstrating that LLMs can generalize tasks from a small number of examples embedded in the input context. 
Min et al.\cite{min-etal-2022-metaicl} propose Meta-training for In-Context Learning (MetaICL), which significantly enhances ICL capabilities through continuous training on various tasks using demonstrations.
Additionally, the concept of supervised context training\cite{chen-etal-2022-improving} is proposed to bridge the gap between pre-training and downstream ICL tasks. 
LLM refines its prior knowledge through ICL, thereby improving performance across multiple tasks \cite{Krishnamurthy2024CanLL}. 
ICL allows a single model to perform various tasks universally, helping it better align its predictions with the semantic requirements of the prompts.

\subsection{Task Decomposition}
Perez et al.\cite{perez-etal-2020-unsupervised} decompose complex problems into several independent subproblems by the LLMs, and then aggregates the answers to form the final response.
Wang et al.\cite{Wang2022IterativelyPP} address problems by modeling prompts as continuous virtual tokens and iteratively eliciting relevant knowledge from a LLM.
Yang et al.\cite{yang-etal-2022-seqzero} decompose normal questions into a series of subproblems, which are then converted into SQL queries using a rule-based system.
Wu et al.\cite{10.1145/3491102.3517582} introduce the idea of linking LLM steps, where the output of one step becomes the input of the next, and developed an interactive system for users to build and modify these chains. 
Zhou et al.\cite{zhou2023leasttomost} argue that generated subproblems are often interdependent and need to be solved in a specific order, with the answers to some subproblems serving as the foundation for others.
They propose the Least-to-Most Prompting method, which links the problem decomposition process to the solving of subproblems.
Zhang et al.\cite{zhang2024cumulative} propose the Cumulative Reasoning (CR), breaking down complex tasks into smaller manageable steps and utilizing iterative collaboration among three different LLMs to incrementally solve problems.

\subsection{Self Evaluation}
Researchers have proposed automated evaluation methods, such as Sentence-BERT \cite{reimers-gurevych-2019-sentence} and SimCSE \cite{gao-etal-2021-simcse}, to assess the reasoning process. 
However, these methods primarily concentrate on matching individual words and phrases, which limits their ability to fully assess the logical consistency and deeper meaning of the context.
To address these limitations, the feasibility of using LLMs to evaluate their own predictions is becoming an increasingly important step in problem-solving. 
Shinn et al.\cite{Shinn2023ReflexionLA}, Kim et al.\cite{10.5555/3666122.3668141} and Paul et al.\cite{paul-etal-2024-refiner} introduce the Self Evaluation (SE) mechanism, where LLMs provide feedback on the candidate answers they generate. 
Chen et al.\cite{chen2024teaching} improve LLM code generation accuracy by using self-generated feedback.
Similarly, Kim et al.\cite{10.5555/3666122.3667845} introduce a review step to evaluate actions and states in operational tasks and decide the next steps.
In terms of reasoning, Kim et al.\cite{NEURIPS2023_271db992} emphasize SE guided decoding, where the LLM uses carefully designed prompts to evaluate candidate answers via a tree search procedure.
Kumar et al.\cite{10.1145/3657604.3662042} explore facilitate scalable self-reflection in LLM, demonstrating its effectiveness in improving student learning outcomes. 
By incorporating fair assessment in LLM learning, our approach injects the reflection mechanism into problem space understanding rather than just evaluating candidate answers, allowing for deeper consideration of the problem and focusing more on the essence of the problem.

\section{Proposed Approach}

\begin{figure*}
	\centering
	\includegraphics[width=\textwidth]{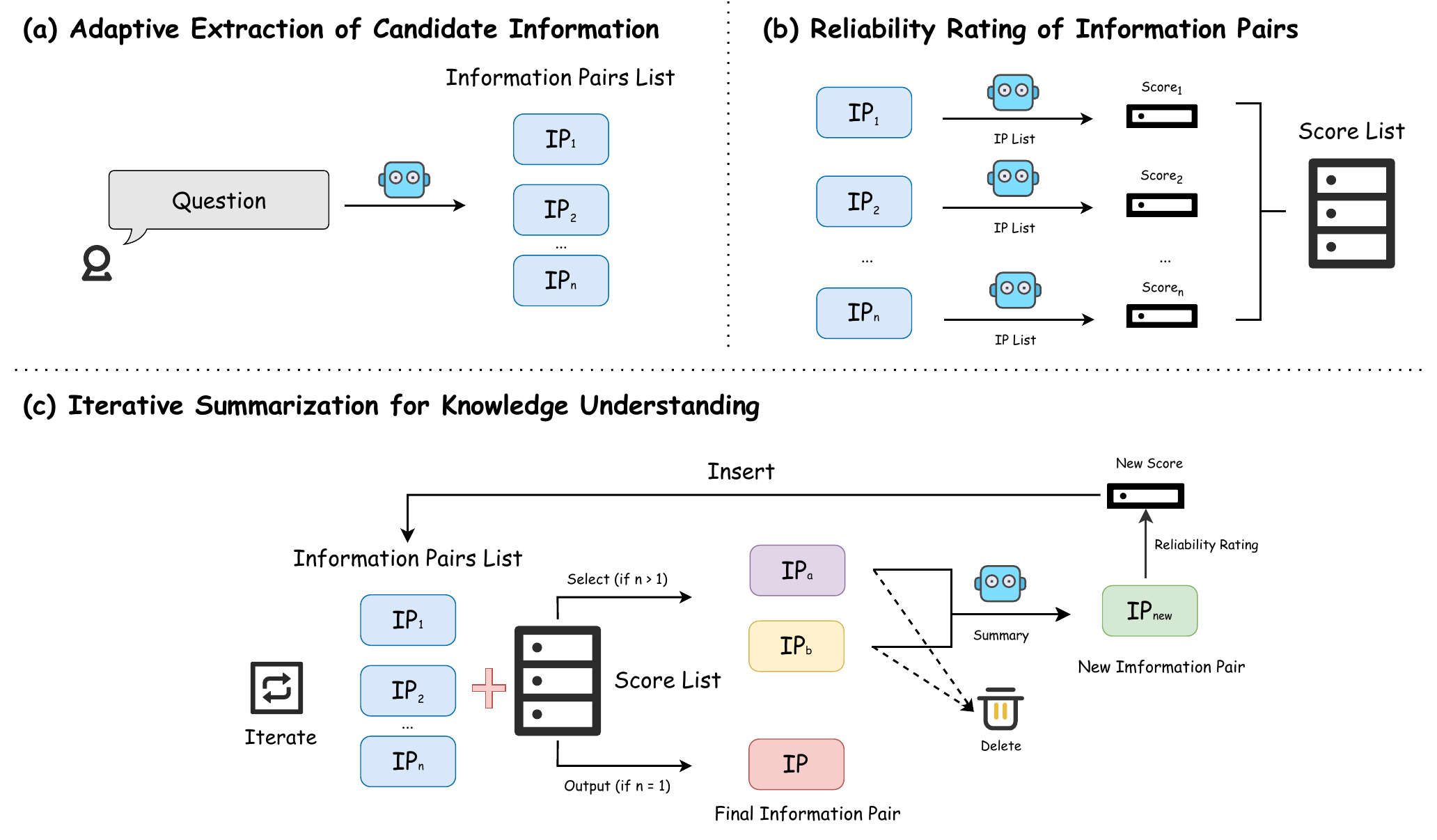} 
	\caption{An illustrative example of the Iterative Summarization Pre-Prompting ($\text{ISP}^{2}$) workflow. $\text{ISP}^{2}$ starts by obtaining an initial set of explicit information pairs based on the current question. It then iteratively refines the description of the problem space through a reliability scoring mechanism and iterative summarization, ultimately arriving at the final answer with the help of fundamental prompts.}
	\label{fig1}
\end{figure*}
The key to problem-solving depends on rigorous semantic parsing of task requirements and systematic integration of contextual information. 
The process requires both precise identification of the key components that contain the answers and formal representation of the problem space through structured knowledge. 
Such formalization reduces the cognitive complexity of problem-solving while explicitly constructing the problem space.
As illustrated in Figure \ref{fig1}, we propose Iterative Summarization Pre-Prompting ($\text{ISP}^{2}$), a plug-and-play pre-prompting method that operates in three steps: adaptive extraction of candidate information, reliability rating of information pairs, and iterative summarization for knowledge understanding.
$\text{ISP}^{2}$ enables continuous optimization of knowledge representations through successive approximation, thereby enhancing the model's capacity for problem resolution.

\subsection{Adaptive Extraction of Candidate Information}

LLMs demonstrate distinct advantages over traditional machine learning methods in extracting and synthesizing complex information, thanks to their sophisticated architectures and extensive training on broad datasets \cite{10.1007/978-3-031-36004-6_69}. 
Given the strong information extraction capabilities, we leverage the LLM to extract multiple entities \( E \) under the question \( Q \).
To ensure extraction relevance, the prompt adopts example-driven learning, providing two template extraction cases in the prompt. 
The template instructions are accompanied by demonstrations of entity extraction for questions in the relevant domain.  
It enables the LLM to learn the relevance criteria for extraction across examples (such as semantic proximity and causal influence), ultimately mimicking the demonstrated behavior to extract the final entity set \( E = \{ e_i \}_{i=1}^{n} \) without requiring an explicit scoring stage.
For each retained entity, the LLM also generates structured information descriptions \( K = \{k_{i1}, k_{i2}, \ldots, k_{it}\}_{i=1}^n \) through a template-driven process. 
These descriptions typically include definitional attributes (e.g., physical laws), contextual constraints (e.g., applicable domains), and operational relationships (e.g., mathematical dependencies).
The entities and their corresponding information descriptions are organized into information pairs \( IP = \left[e_i,\{k_{i1}, k_{i2}, \ldots, k_{it}\}\right]_{i=1}^n \). 
This step of acquiring information pairs is called adaptive extraction, where each information pair focuses on summarizing the explicit information in the problem.
Adaptive extraction lays the foundation for subsequent iterative summarization, where explicit information serves as an anchor for inferring implicit information through structured knowledge propagation.
Meanwhile, we do not specify the number of entities or information descriptions due to the varying amount of extracted information for each question. 
This flexible approach ensures that the LLM can adapt to the specific requirements of different problems, continuously enhancing its understanding and refining its outputs.
The entity and description prompts are structured as follows:
\begin{center}
	\begin{tcolorbox}[colback=blue!5!white,colframe=blue!55!black,width=3.5in,title={Entity Prompting}]
		{
			{
				\texttt{\{2 Entity Template\}}\\
				\texttt{Q: \{Input Query\}} \\
				\texttt{Generate key entities:} \\
				\texttt{\#{ Thought-eliciting prompt}  (e.g.,``Let’s think step by step") \#}
			}
		}
	\end{tcolorbox}
\end{center}
\begin{center}
	\begin{tcolorbox}[colback=blue!5!white,colframe=blue!55!black,width=3.5in,title={Description Prompting}]
		{
			{
				\texttt{\{2 Description Template\}}\\
				\texttt{Q: \{Input Query and Entities\}} \\
				\texttt{Generate descriptions:} \\
				\texttt{\#{ Thought-eliciting prompt}  (e.g.,``Let’s think step by step") \#}
			}
		}
	\end{tcolorbox}
\end{center}

\subsection{Reliability Rating of Information Pairs}

Inspired by the recent success of Self Evaluation \cite{Kadavath2022LanguageM}, we introduce an automated evaluation method called reliability rating.
It quantifies the quality and reliability of information pairs, assessing their problem-solving potential and completeness.
It evaluates the current information pair $IP_t$ conditioned on all preceding pairs $IP_{1:n}$.  
By leveraging a LLM, we compute the reliability rating as an expectation over the model's probability distribution:
\begin{equation}
	V(IP_t) = \mathbb{E}_{v \sim p_{\text{LLM}}(v \mid T, Q, IP_{1:n})} [ v ],
\end{equation}
where $p_{\text{LLM}}(v \mid \cdot)$ represents the LLM's probabilistic assessment of $IP_t$'s utility.  
Conditioning on $T$, $Q$, and $IP_{1:n}$ allows reliability rating to integrate contextual dependencies into the evaluation process. 
It ensures that the reliability score reflects not merely a single deterministic outcome, but a comprehensive assessment of the information's quality within the problem-solving context.
By anchoring the evaluation in the interplay between task-specific requirements, question, and the evolving information pairs, the reliability rating becomes sensitive to the logical coherence and relevance of information pairs.
The expectation-based evaluation method makes integrate diverse factors contributing to reliability(such as fluency, relevance, and completeness)into a unified score. 
Rather than relying on a rigid weighted sum of individual dimensions, this method leverages the LLM's learned representations to balance these factors dynamically, resulting in a more flexible and interpretable metric.  
LLMs inherently capture the uncertainty and variability in natural language generation, and their probability distributions provide a principled way to quantify the likelihood of different outputs.
In scenarios where exact computation of the expectation $\mathbb{E}_{v \sim p_{\text{LLM}}(v \mid T, Q, IP_{1:n})} [v]$ is infeasible or computationally expensive, we approximate it using multiple sampling. 
Specifically, we draw $k$ independent samples $v^{(i)}$ from the conditional probability distribution $p_{\text{LLM}}(v \mid T, Q, IP_{1:n})$ and compute the reliability rating as the empirical average:
\begin{equation}
	V(IP_t) = \frac{1}{k} \sum_{i=1}^k v^{(i)}, \quad v^{(i)} \sim p_{\text{LLM}}(v \mid T, Q, IP_{1:n}),
\end{equation}
where $v^{(i)}$ represents the score assigned to the $i$-th sample. 
The number of samples $k$ controls the trade-off between estimation accuracy and computational cost. Given the experimental constraints, we set the number of samples to 3.

The prompts \( T \) for evaluating information pairs are divided into two types:
\begin{itemize}
	\item \textbf{Scalar Value Prompt}: Directly prompts the LLM to output a scalar value \( v \) (ranging from 0 to 1).
	\item \textbf{Opinion-Based Judgement Prompt}:  Prompts the LLM to generate opinion-based judgments (e.g., absolutely reliable, moderately reliable, weakly reliable, unreliable), which can be converted into numerical values \( v \) (1, 0.67, 0.33, 0).
\end{itemize}
To construct the prompt \( T \), we provide stepwise evaluation examples (similar to question answering with rationales) for each instance. 
Reliability rating prompt takes different forms depending on the specific problem, enabling the LLM to assess the information pair and assign an appropriate value based on its judgment. 
In mathematical tasks, the reliability rating places greater emphasis on logical reasoning and the correctness of derivations, as these are critical for ensuring accurate solutions. 
In contrast, for commonsense reasoning or general knowledge tasks, the focus shifts toward coherence, relevance, and the ability to provide intuitive and contextually appropriate responses.

\subsection{Iterative Summarization for Knowledge Understanding}

\begin{figure}
	\centering
	\includegraphics[width=3.5in]{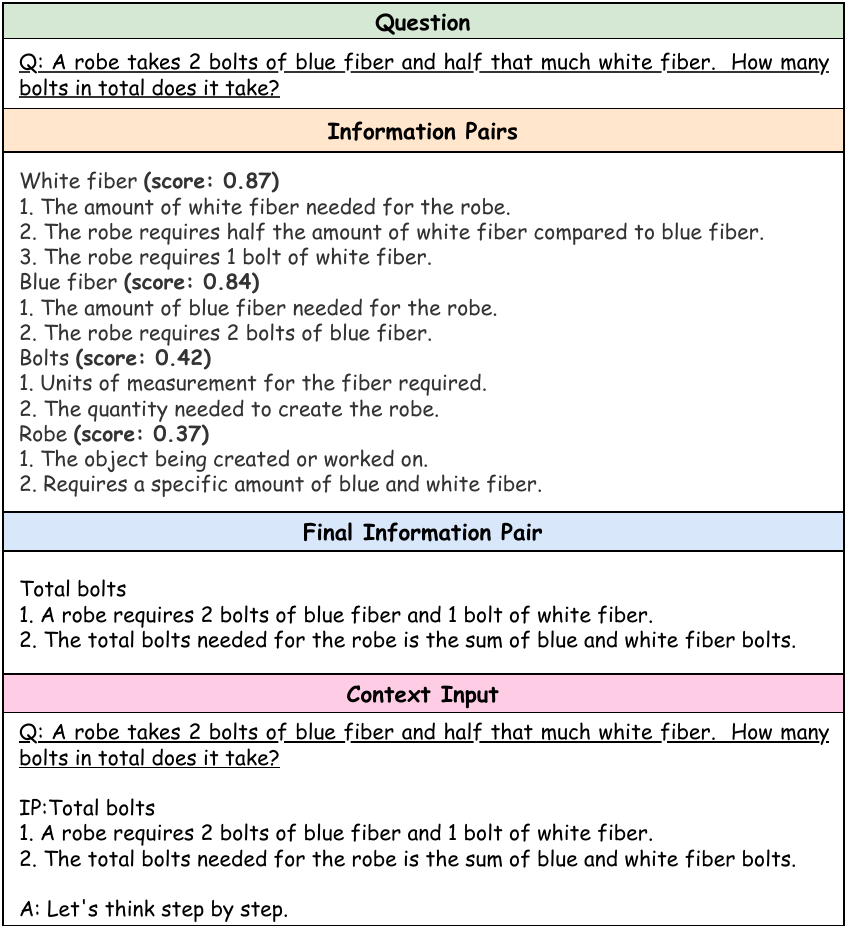}
	\caption{Example inputs of CoT prompting with $\text{ISP}^{2}$.}
	\label{fig2}
\end{figure}

To transform fragmented information into deeply integrated knowledge, we propose an iterative approach that adaptively summarizes and refines the extracted information pairs, progressively distilling knowledge that ultimately contributes to effective problem-solving.
It operates through three recurrent operations: selection of two information pairs with minimal reliability scores, generative synthesis of composite knowledge, and reliability rating of the new information pair.

Figure \ref{fig1} illustrates the computational pipeline, while Figure \ref{fig2} demonstrates its application in a mathematical dataset.
The core of the iterative process lies in summarizing through the merging of two low-reliability information pairs.
Among the given $ n $ information pairs, the two with the lowest reliability scores are selected for merging in each iteration, rather than those with higher scores.
The transition from low to high reliability scores reflects the evolution of knowledge from incomplete to complete information, representing a process of deepening understanding. 
For example, in the initial stage, the algorithm may select the information pairs \textit{"Bolts"} (score: 0.42) and \textit{"Robe"} (score: 0.37). 
Unlike traditional methods that favor high scores inputs, this strategy leverages the insight that low-reliability pairs often contain complementary fragments of knowledge. 
Merging such pairs can activate latent relationships and enable implicit inference. 
Combining the pairs \textless Bolts, ``1. Units of measurement for the fiber required. 2. The quantity needed to create the robe'' \textgreater{} and \textless Robe, ``1. Requires a specific amount of blue and white fiber'' \textgreater{}, a tailored prompt guides the LLM to generate a new information pair: \textless Total bolts, ``1. A robe requires 2 bolts of blue fiber and 1 bolt of white fiber. 2. The total bolts needed for the robe is the sum of blue and white fiber bolts'' \textgreater{}. 
The reliability score of the new pair increases to 0.78, indicating the successful transformation from explicit data to refined implicit knowledge.
The merging operation follows a probabilistic maximization principle, where the new information pair $IP_{\text{new}}$ is defined as:
\begin{equation}
	IP_{\text{new}} = \arg\max_{IP \subseteq IP_a \cup IP_b} \prod_{k=1}^m p_k(IP \mid x)
\end{equation}
where $p_k$ denotes the contextual probability of the $k$-th knowledge element given the context $x$. 
After the merge, the new information pair replaces the original two pairs $IP_a$ and $IP_b$, and the reliability scores of the entire list are recalculated. 
The process continues iteratively until it meets the following termination criteria: the list converges to a single information pair, regardless of its reliability score. 
It ensures termination either when no further merging is possible.

\subsection{Question Answering}

\begin{center}
	\begin{tcolorbox}[colback=blue!5!white,colframe=blue!55!black,width=3.5in,title={ISP$^2$-CoT Reasoning}]
		\textbf{Problem:} A person travels at 20 km/h and reaches the destination in 2.5 hours. Find the distance.\\
		\textbf{Options:} A) 53 km,\quad B) 55 km,\quad C) 52 km,\quad D) 60 km,\quad E) 50 km
		
		\vspace{0.8em}
		\textbf{Information Pair}\\
		\texttt{Velocity:}\\
		\texttt{1. The person is traveling at a constant speed of 20 km/hr.} \\
		\texttt{2. The travel time to reach the destination is 2.5 hours.} \\
		\texttt{3. The distance traveled can be found by multiplying the speed by the travel time.} \\

		\vspace{0.8em}
		\textbf{Let's think step by step.}\\
		\texttt{The distance that the person traveled would have been 20 km/hr * 2.5 hrs = 50 km.} 
		
		\vspace{0.8em}
		\textbf{Answer:} \fbox{\textbf{E}} (50 km)
	\end{tcolorbox}
\end{center}

We treat the reasoning process of LLMs as an autoregressive generation task. 
Typically, the input context $x$ consists of two parts: the question $Q$ and the prompt $T$. Given $x$, the model needs to generate the final result $y$.
To generate a reasonable $y$, the LLM needs to leverage CoT and reason correctly through $z$ as an intermediate step. 
We define the predictive probability formula as follows:
\begin{equation}
	p(y \mid x=(T, Q)) = p(y \mid x, z) \cdot p(z \mid x),
\end{equation}
where $p(y \mid x, z)$ represents the probability of generating the final result $y$ given both the input context $x$ and the intermediate reasoning step $z$, and $p(z \mid x)$ represents the probability of generating the intermediate reasoning step $z$ given the input context $x$.
$y$ and $z$ are conditionally independent given $x$. 
The intermediate reasoning step $z$ can be viewed as independently generated from the final result $y$, conditioned on the input $x$. 
In practice, this conditional independence reflects the modular nature of the reasoning process, where each step contributes to the overall generation without direct dependence on subsequent steps.

Building on the foundation, ISP$^2$ enhances reasoning capabilities by integrating information extraction with cognitive summarization. 
It employs an iterative summarization process before inference, progressively refining information pairs based on the given question. 
ISP$^2$ not only extracts relevant information but also deeply comprehends the context and nuances of the problem at hand.
The method allows the LLM to generate more accurate and contextually relevant information pairs for the final reasoning step. Ultimately, the unique information pair IP is combined with the question for the final reasoning process. Formally, it can be represented as:
\begin{equation}
	p(y \mid x=(T, Q, IP)) = p(y \mid x, z) \cdot p(z \mid x),
\end{equation}
where IP represents the unique information pair generated through iterative refinement.

To further improve the quality and accuracy of reasoning results, ISP$^2$ is designed to be compatible with different CoT prompting methods. 
Whether the task requires simple step-by-step reasoning or more sophisticated in-context learning, ISP$^2$ dynamically integrates the appropriate prompting strategy to optimize performance.
We illustrate an example of ISP$^2$ reasoning in the figure above.

\section{Experiments}

In this section, we present a comprehensive overview of the experiments conducted to evaluate the performance and effectiveness of our proposed method. 
The experimental setup includes detailed descriptions of the datasets used, evaluation metrics, and baseline models. 
The main results are presented in subsequent subsections. Additionally, we analyze the performance across different steps and highlight key findings.

\subsection{Experimental Setup}

\paragraph{Tasks and Datasets}
We evaluate $\text{ISP}^{2}$ on six datasets with diverse input formats. 
Extensive experiments are conducted across these datasets to demonstrate the universality of $\text{ISP}^{2}$ prompts.
The Table \ref{tab:dataset-statistic} provides relevant information about the datasets used in our experiment, detailing the data source, task type, answer type, number of prompt samples, and total test samples for each dataset.
\begin{itemize}
	\item \textbf{Mathematical Reasoning:} The following four mathematical reasoning benchmarks are widely recognized and considered in the field: AddSub \cite{hosseini-etal-2014-learning}, which includes math word problems on addition and subtraction tailored for third to fifth graders; SVAMP \cite{patel-etal-2021-nlp}, known for its math word problems with diverse structures; AQuA \cite{ling-etal-2017-program}, which focuses on algebraic word problems; GSM8K \cite{Cobbe2021TrainingVT}, a published benchmark that features grade-school math problems.
	\item \textbf{Commonsense Reasoning:} StrategyQA (SQA; \cite{10.1162/tacl_a_00370}) and CommonsenseQA (CSQA; \cite{talmor-etal-2019-commonsenseqa}) are utilized for commonsense tasks. CSQA comprises questions that require a variety of commonsense knowledge, while StrategyQA includes questions that necessitate multi-step reasoning.
\end{itemize}

\begin{table}[!b]
	\centering
	\caption{Overview of datasets utilized in experiments}
	\label{tab:dataset-statistic}
	\resizebox{3.5in}{!}{
		\begin{tabular}{|l|c|c|c|c|}
			\hline
			Dataset & Reasoning Task & Answer Type &  Example & Number  \\
			\hline
			\href{https://github.com/openai/grade-school-math}{GSM8K} \cite{Cobbe2021TrainingVT} & Arithmetic & Number & 4 & 1,319  \\
			\href{https://github.com/wangxr14/Algebraic-Word-Problem-Solver}{AddSub} \cite{hosseini-etal-2014-learning} & Arithmetic & Number & 4 & 395  \\
			\href{https://github.com/arkilpatel/SVAMP}{SVAMP} \cite{patel-etal-2021-nlp}& Arithmetic & Number & 4 & 1,000  \\
			\href{https://github.com/deepmind/AQuA}{AQuA} \cite{ling-etal-2017-program}& Arithmetic & Multi-choice & 4 & 254  \\
			
			\href{https://github.com/eladsegal/strategyqa}{StrategyQA} \cite{10.1162/tacl_a_00370}& Commonsense & True/False & 4 & 2,290 \\
			\href{https://www.tau-nlp.sites.tau.ac.il/commonsenseqa}{CommonsenseQA} \cite{talmor-etal-2019-commonsenseqa} & Commonsense & Multi-choice & 4 & 1,221  \\
			\hline
		\end{tabular}
	}
	
	\vspace{2mm}
	\footnotesize Note: ``Number'' represents the number of sampled datasets, and ``Example'' is the number of prompt examples in the same dataset.
\end{table}

\paragraph{Base Prompting}
To effectively evaluate our method, we assess $\text{ISP}^{2}$ performance on three baseline prompting methods: Chain-of-Thoughts (CoT) \cite{wei2022chain}, Complex CoT \cite{fu2023complexitybased}, Program-of-Thoughts (PoT) \cite{chen2023program} and Self Consistency \cite{wang2023selfconsistency}. 
CoT predicts answers by generating explanations and steps, allowing the model to solve problems through explicit reasoning processes, which makes the decision-making more transparent. 
Complex CoT utilizes a complexity-based strategy, which breaks down complex problems into smaller parts, thereby improving reliability in complex scenarios. 
PoT prompts LLMs to generate executable code for problem decomposition and delegates computational steps to external interpreters, thereby disentangling reasoning from numerical processing to enhance accuracy and efficiency in complex tasks.
Self Consistency generates multiple chains of thought and selects the highest-voted result as the final outcome through voting.
Additionally, all experiments are conducted in a few-shot setting without training or fine-tuning the LLMs.

\paragraph{LLMs and Implementations}

We primarily use GPT-3.5, GPT-4o-mini, LLaMA2-7B, LLaMA2-13B, LLaMA3-8B, and DeepSeek-Distill-Qwen-7B for testing.
Our decoding strategy employs \textit{greedy decoding} with the temperature fixed at 0, ensuring deterministic outputs to eliminate stochasticity in comparisons between baseline methods and ISP$^2$-enhanced variants. 
It minimizes the risk of hallucinations (e.g., generating inconsistent or nonsensical reasoning steps) while aligning with our experimental goal of isolating the intrinsic impact of ISP$^2$. 
By fixing the temperature at 0, we ensure comparability across all methods and avoid conflating the effects of ISP$^2$ with those of decoding stochasticity.

For the few-shot setting, we use four exemplars for most datasets, adjusting the number based on task complexity. 
We integrate ISP$^2$ into baseline methods to evaluate its impact, denoted as ISP$^2$-CoT, ISP$^2$-CoT@5, ISP$^2$-ComCoT, ISP$^2$-ComCoT@5, ISP$^2$-PoT, and ISP$^2$-PoT@5. 
Here, ISP$^2$-CoT, ISP$^2$-ComCoT, and ISP$^2$-PoT represent ISP$^2$ combined with Chain-of-Thought (CoT), Complex CoT, and Program-of-Thoughts (PoT), respectively. The ``@5'' notation indicates the use of Self-Consistency (SC) by aggregating five reasoning chains via majority voting. 
Despite temperature=0, SC@5 introduces diversity through parallel sampling of distinct reasoning paths (enabled by varying the random seed in API calls), compensating for the reduced per-sample stochasticity.
Additionally, we design task-specific answer format instructions in prompts to regulate the structure of final outputs (e.g., ``Answer: [X]'') for precise answer extraction.
Notably, the \textbf{ISP\textsuperscript{2}-PoT} method specifically selects \textbf{LLMs} with strong code-generation capabilities (e.g., GPT and DeepSeek). By decoupling program derivation from code execution, it avoids the problem of solution failure caused by the model's insufficient code implementation ability.

\subsection{Main Results}

\begin{table}
	\centering
	\caption{The comparison results of existing SOTA plug-and-play methods on GPT-3.5.}
	\label{tab:performance_scores}
	\resizebox{3.5in}{!}{
		\begin{tabular}{|c|c|cccccc|c|}
			\hline
			\multirow{7}{*}{\shortstack{GPT-3.5 \\ turbo}} &Prompt & AddSub & SVAMP & GSM8K & AQuA & CSQA & SQA & Average \\ 
			\hline
			&CoT~\cite{wei2022chain}    & 81.2 & 81.2 & 76.2 & 58.7 & 75.5 & 69.7 & 73.75 \\ 
			&PHP-CoT~\cite{zheng2024progressivehint} & \underline{86.1} & 83.1 &\textbf{84.6} &\textbf{65.4} & 76.2 & 69.2 & \underline{77.43} \\ 
			&RE2-CoT~\cite{xu2024rereading}  & 82.7 & \underline{84.9} & 81.2 & 63.3 & 77.9 & 67.1 & 76.60 \\ 
			&ERA-CoT~\cite{liu-etal-2024-era}  & 83.8 &82.2  & 80.2 & 56.9 & \textbf{83.2} & \textbf{71.4} & 76.28 \\ 
			&$\text{ISP}^{2}$-CoT   &\textbf{88.6}  &\textbf{88.7}   &\underline{83.4}  &\underline{64.4} & \underline{78.7} & \textbf{71.4} & \textbf{79.43} \\ 
			\hline
		\end{tabular}
	}
	
	\vspace{2mm}
	\footnotesize Note: We use accuracy as the evaluation metric.The best result is highlighted in \textbf{bold}, and the second best is \underline{underlined}.
\end{table}

The main results are presented in Table \ref{experiment_2}. 
Compared with the SOTA method that also functions as a plug-in, $\text{ISP}^{2}$ demonstrates considerable advantages. 
Significant improvements have been achieved on benchmarks such for GPT-4o-mini, GPT-3.5, Deepseek-Qwen-7B, LLaMA3-8B, LLaMA2-13B, and LLaMA2-7B. The average improvement rates were 1.1\% for GPT-4o-mini, 7.1\% for GPT-3.5, 4.5\% for DeepSeek-Qwen-7B, 5.2\% for LLaMA3-8B, 8.1\% for LLaMA2-13B, and 12.4\% for LLaMA2-7B.
These results indicate that by processing $\text{ISP}^{2}$, LLM can better understand the essence of the problems and enhance its performance.

\begin{table}
	\centering
	\caption{Result on Math Word Problem}
	\label{experiment_2}
	\resizebox{3.4in}{!}{
	\begin{tabular}{|c|c|cccc|c|}
		\hline
		\multirow{2}{*}{Model} & \multirow{2}{*}{Method} & \multicolumn{4}{c|}{Dataset} & \multirow{2}{*}{Average}\\
		\cline{3-6}
		& & AddSub & SVAMP & GSM8K &AQuA &\\
		\hline
		\multirow{18}{*}{GPT-4o-mini}  
		& CoT & 96.1 & 93.4 & 92.3 & 81.4 &90.80 \\
		&\textbf{$\text{ISP}^{2}$-CoT} & 97.2 & 93.9 & 92.4 & 83.7 &91.80\\
		& & \textcolor{red}{+1.1} & \textcolor{red}{+0.5} & \textcolor{red}{+0.1} & \textcolor{red}{+2.3} & \textcolor{red}{+1.00}\\
		\cline{2-7}
		&PoT & 92.1 & 91.4 & 91.4 & 81.1 & 89.00 \\
		&\textbf{$\text{ISP}^{2}$-PoT} & 93.7 & 92.1 & 92.1 & 81.7 & 89.90 \\
		& & \textcolor{red}{+1.6} & \textcolor{red}{+0.7} & \textcolor{red}{+0.7} & \textcolor{red}{+0.6} & \textcolor{red}{+0.90} \\ 
		\cline{2-7} 
		&ComCoT &96.7  &93.8  &91.8  &82.3  &91.15  \\
		&\textbf{$\text{ISP}^{2}$-ComCoT} &97.2  &94.1  &92.3  &83.7 &91.83  \\
		& & \textcolor{red}{+0.5} & \textcolor{red}{+0.3} & \textcolor{red}{+0.5} & \textcolor{red}{+1.4} & \textcolor{red}{+0.68} \\
		\cline{2-7}
		&CoT@5 & 96.3 & 93.5 & 93.3 & 83.1 &91.55 \\
		&\textbf{$\text{ISP}^{2}$-CoT@5} & 98.5 & 93.9 & 94.3 & 84.2 &92.73 \\
		& & \textcolor{red}{+2.2} & \textcolor{red}{+0.4} & \textcolor{red}{+1.0} & \textcolor{red}{+0.9} & \textcolor{red}{+1.18}\\
		\cline{2-7}
		&PoT@5 & 95.6 & 94.1 & 92.2 & 81.7 & 90.90 \\
		&\textbf{$\text{ISP}^{2}$-PoT@5} & 97.0 & 94.3 & 92.8 & 82.7 & 91.70 \\
		& & \textcolor{red}{+1.4} & \textcolor{red}{+0.2} & \textcolor{red}{+0.6} & \textcolor{red}{+1.0} & \textcolor{red}{+0.80} \\
		\cline{2-7}
		&ComCoT@5 &97.1  &93.9  &92.6  &83.0  &91.65  \\
		&\textbf{$\text{ISP}^{2}$-ComCoT@5} &97.9  &94.2  &92.8  &86.9  &92.95  \\
		& & \textcolor{red}{+0.8} & \textcolor{red}{+0.3} & \textcolor{red}{+0.2} & \textcolor{red}{+3.9} & \textcolor{red}{+1.30} \\
		\hline
		\multirow{18}{*}{GPT-3.5 Turbo}   
		& CoT & 81.2 & 81.2 & 76.2 & 58.7 & 74.33\\
		&\textbf{$\text{ISP}^{2}$-CoT} & 88.6 & 88.7 & 83.4 & 64.4 & 81.28\\
		& & \textcolor{red}{+7.4} & \textcolor{red}{+7.5} & \textcolor{red}{+7.2} & \textcolor{red}{+5.7} & \textcolor{red}{+6.95}\\
		\cline{2-7}
		&PoT &84.4  &85.2  &81.8  &58.1  &77.38  \\
		&\textbf{$\text{ISP}^{2}$-PoT} &90.1  &87.7  &83.2  &66.3  &81.83  \\
		& & \textcolor{red}{+5.7} & \textcolor{red}{+2.5} & \textcolor{red}{+1.4} & \textcolor{red}{+8.2} & \textcolor{red}{+4.45} \\ 
		\cline{2-7}
		&ComCoT & 82.7 & 80.1 & 79.3 & 57.8 & 74.98 \\
		&\textbf{$\text{ISP}^{2}$-ComCoT} & 89.1 & 87.2 & 84.6 & 63.7 & 81.15 \\
		& & \textcolor{red}{+6.4} & \textcolor{red}{+6.4} & \textcolor{red}{+5.3} & \textcolor{red}{+5.9} & \textcolor{red}{+6.17} \\
		\cline{2-7}
		&CoT@5 & 82.3 & 85.2 & 80.8 & 66.5 & 78.70\\
		&\textbf{$\text{ISP}^{2}$-CoT@5} & 93.9 & 90.1 & 84.8 & 72.7 & 85.38\\
		& & \textcolor{red}{+11.6} & \textcolor{red}{+4.9} & \textcolor{red}{+4.0} & \textcolor{red}{+6.2} & \textcolor{red}{+6.68}\\
		\cline{2-7}
		&PoT@5 &86.8  &86.6  &82.7  &58.6  &78.68  \\
		&\textbf{$\text{ISP}^{2}$-PoT@5} &91.7  &90.8  & 84.9 & 72.2 &84.90  \\
		& & \textcolor{red}{+4.9} & \textcolor{red}{+8.1} & \textcolor{red}{+2.2} & \textcolor{red}{+13.6} & \textcolor{red}{+6.22} \\
		\cline{2-7}
		&ComCoT@5 & 83.9 & 84.4 & 83.9 & 64.6 & 78.98 \\
		&\textbf{$\text{ISP}^{2}$-ComCoT@5} & 92.8 & 90.8 & 87.0 & 70.5 & 85.28 \\
		& & \textcolor{red}{+8.9} & \textcolor{red}{+6.4} & \textcolor{red}{+3.1} & \textcolor{red}{+5.9} & \textcolor{red}{+6.30} \\
		\hline
		\multirow{18}{*}{\makecell{DeepSeek-\\Distill-\\Qwen-7B}} 
		& CoT &89.1  &86.9  &75.3  &47.2 &74.63 \\
		&\textbf{$\text{ISP}^{2}$-CoT} &90.5  &91.4  &82.8  &48.9 &78.40 \\
		& & \textcolor{red}{+1.4} & \textcolor{red}{+4.5} & \textcolor{red}{+7.5} & \textcolor{red}{+1.7} & \textcolor{red}{+3.77}\\
		\cline{2-7}
		&PoT &83.5 & 77.4 & 70.5 &30.2  &65.40  \\
		&\textbf{$\text{ISP}^{2}$-PoT} &88.7  &77.9  & 78.2 &32.1  &69.23  \\
		& & \textcolor{red}{+5.2} & \textcolor{red}{+0.5} & \textcolor{red}{+7.7} & \textcolor{red}{+1.9} & \textcolor{red}{+3.83} \\
		\cline{2-7}
		&ComCoT & 88.4 &93.4  &78.2  &62.6  &80.65  \\
		&\textbf{$\text{ISP}^{2}$-ComCoT} &91.5  &94.3  &86.3  &65.8  &84.48  \\
		& & \textcolor{red}{+3.1} & \textcolor{red}{+0.9} & \textcolor{red}{+8.1} & \textcolor{red}{+3.2} & \textcolor{red}{+3.83} \\
		\cline{2-7}
		&CoT@5 &90.6  &87.8  &77.2  &53.7 &77.33 \\
		&\textbf{$\text{ISP}^{2}$-CoT@5}  &92.2  &91.8  &83.1  &56.5 &80.90 \\
		& & \textcolor{red}{+1.6} & \textcolor{red}{+4.0} & \textcolor{red}{+5.9} & \textcolor{red}{+2.8} & \textcolor{red}{+3.57}\\
		\cline{2-7}
		&PoT@5 & 84.4 & 80.0 & 71.8 &32.3  &67.13  \\
		&\textbf{$\text{ISP}^{2}$-PoT@5} &92.5  &82.5  & 79.8 &33.3  &69.78  \\
		& & \textcolor{red}{+8.1} & \textcolor{red}{+2.5} & \textcolor{red}{+8.0} & \textcolor{red}{+1.0} & \textcolor{red}{+2.65} \\
		\cline{2-7}
		&ComCoT@5 &90.6  &95.1  &79.8  &67.3  & 83.20 \\
		&\textbf{$\text{ISP}^{2}$-ComCoT@5} &93.4  &95.6  & 87.2 &67.9  &86.02  \\
		& & \textcolor{red}{+2.8} & \textcolor{red}{+0.4} & \textcolor{red}{+9.9} & \textcolor{red}{+0.6} & \textcolor{red}{+2.82} \\
		\hline
		\multirow{12}{*}{LLaMA3-8B}  
		& CoT &75.3  &66.8  &45.9  &22.7  &52.68 \\
		&\textbf{$\text{ISP}^{2}$-CoT} & 75.6 & 67.9 & 49.1 & 23.8 & 54.10 \\
		& & \textcolor{red}{+0.3} & \textcolor{red}{+1.1} & \textcolor{red}{+3.2} & \textcolor{red}{+1.1} & \textcolor{red}{+1.42}\\  
		\cline{2-7}
		&ComCoT &75.6  &66.2  &49.7  &25.0  &54.13  \\
		&\textbf{$\text{ISP}^{2}$-ComCoT} &76.5  &73.0  &50.6  &27.8  &56.98  \\
		& & \textcolor{red}{+0.9} & \textcolor{red}{+6.8} & \textcolor{red}{+0.9} & \textcolor{red}{+2.8} & \textcolor{red}{+2.85} \\
		\cline{2-7}
		&CoT@5 &76.7  &69.4  &47.6  &24.1 &54.45 \\
		&\textbf{$\text{ISP}^{2}$-CoT@5} &77.8  &69.6  &49.3  &26.4 &55.78 \\
		& & \textcolor{red}{+1.1} & \textcolor{red}{+0.2} & \textcolor{red}{+1.7} & \textcolor{red}{+2.3} & \textcolor{red}{+1.33}\\
		\cline{2-7}
		&ComCoT@5 &77.1  &66.4  &50.3  &26.4  &55.05  \\
		&\textbf{$\text{ISP}^{2}$-ComCoT@5} &77.5 &74.8  &51.7  &27.9  &57.98  \\
		& & \textcolor{red}{+0.4} & \textcolor{red}{+8.4} & \textcolor{red}{+1.4} & \textcolor{red}{+1.5} & \textcolor{red}{+2.93} \\
		\hline
		\multirow{12}{*}{LLaMA2-13B}    
		& CoT & 38.1 & 36.3 & 16.7 & 16.6 & 26.93\\
		&\textbf{$\text{ISP}^{2}$-CoT} & 60.0 & 39.2 & 19.3 & 20.4 & 34.73\\
		& & \textcolor{red}{+21.9} & \textcolor{red}{+2.9} & \textcolor{red}{+2.6} & \textcolor{red}{+3.8} & \textcolor{red}{+7.80}\\
		\cline{2-7}
		&ComCoT &32.2 & 33.9 & 15.5 & 19.7 & 25.33 \\
		&\textbf{$\text{ISP}^{2}$-ComCoT} & 70.9 & 47.8 & 18.8 & 20.4 & 39.48 \\
		& & \textcolor{red}{+38.7} & \textcolor{red}{+13.9} & \textcolor{red}{+3.3} & \textcolor{red}{+0.7} & \textcolor{red}{+14.15} \\
		\cline{2-7}
		&CoT@5 & 47.8 & 47.8 & 17.8 & 24.4 &30.45\\
		&\textbf{$\text{ISP}^{2}$-CoT@5} & 64.6 & 49.7 & 21.9 & 25.4 & 40.40 \\
		& & \textcolor{red}{+16.8} & \textcolor{red}{+1.9} & \textcolor{red}{+4.1} & \textcolor{red}{+1.0} & \textcolor{red}{+9.95}\\
		\cline{2-7}
		&ComCoT@5 & 52.1 & 43.6 & 16.9 & 23.6 & 34.05 \\
		&\textbf{$\text{ISP}^{2}$-ComCoT@5} & 74.9 & 57.1 & 20.4 & 24.4 & 44.20 \\
		& & \textcolor{red}{+22.8} & \textcolor{red}{+13.5} & \textcolor{red}{+3.5} & \textcolor{red}{+0.8} & \textcolor{red}{+10.15} \\
		\hline
		\multirow{12}{*}{LLaMA2-7B}   
		& CoT & 29.4 & 30.7 & 7.4 & 19.7 & 21.80\\
		&\textbf{$\text{ISP}^{2}$-CoT} & 35.2 & 39.1 & 8.3 & 21.6 & 26.05\\
		& & \textcolor{red}{+5.8} & \textcolor{red}{+8.4} & \textcolor{red}{+0.9} & \textcolor{red}{+1.9} & \textcolor{red}{+4.25}\\ 
		\cline{2-7}
		&ComCoT & 27.8 & 30.7 & 8.4 & 22.1 & 22.25 \\
		&\textbf{$\text{ISP}^{2}$-ComCoT} & 36.8 & 38.3 & 8.6 & 25.9 & 27.40 \\
		& & \textcolor{red}{+9.0} & \textcolor{red}{+7.6} & \textcolor{red}{+0.2} & \textcolor{red}{+3.8} & \textcolor{red}{+5.15} \\
		\cline{2-7}
		&CoT@5 & 41.8 & 33.4 & 7.8 & 21.6 & 26.15\\
		&\textbf{$\text{ISP}^{2}$-CoT@5} & 43.9 & 42.5 & 11.4 & 29.9 & 31.93\\
		& & \textcolor{red}{+2.1} & \textcolor{red}{+9.1} & \textcolor{red}{+3.6} & \textcolor{red}{+8.3} & \textcolor{red}{+5.78}\\
		\cline{2-7}	
		&ComCoT@5 & 39.9 & 34.2 & 10.5 & 24.5 & 27.28 \\
		&\textbf{$\text{ISP}^{2}$-ComCoT@5} & 43.5 & 43.6 & 12.8 & 27.2 & 31.78 \\
		& & \textcolor{red}{+3.6} & \textcolor{red}{+9.4} & \textcolor{red}{+2.3} & \textcolor{red}{+2.7} & \textcolor{red}{+4.50} \\
		\hline
	\end{tabular}
	}
	
	\vspace{2mm}
	\footnotesize Note: The @5 notation indicates usage of Self-Consistency with five reasoning chains for majority voting. "ComCoT" stands for Complex CoT. "PoT" stands for Program-of-Thoughts.
\end{table}

\paragraph{Mathematical Reasoning} 

Table \ref{experiment_2} reports performance on the Math Word Problem (MWP) task, where our method achieves significant performance improvements across various mathematical subdomains. 
Our method not only achieves performance improvements in standard reasoning approaches, such as CoT, but also enables LLMs to enhance their performance through code generation.
Notably, the enhancement is particularly significant when combined with Self Consistency, and the robustness and accuracy of $\text{ISP}^{2}$ on different models are evident.
The AddSub dataset focuses on basic mathematical operations, and for models like GPT and Llama, they already possess sufficient capabilities to solve most problems. 
However, problems that are not successfully solved usually contain misleading information, and $\text{ISP}^{2}$ can avoid the influence of misleading information by filtering out unnecessary data in adaptive extraction, thus enabling more successful problem solving on LLMs. 
SVAMP and GSM8K are more advanced MWP datasets that focus on generalization capabilities in arithmetic and more complex algebra and geometry problems. 
Interestingly, we observe that the improvements of $\text{ISP}^{2}$ on SVAMP and GSM8K are not less than those on AddSub. 
$\text{ISP}^{2}$ can correct wrong directions through adaptive extraction and prevent misleading information from contaminating the summary process. 
Unlike the first three datasets, AQuA is an algebraic multiple-choice dataset, which not only requires the model to solve the problem but also to select the correct answer from multiple options. 
The format demands higher judgment capabilities from the model. 
$\text{ISP}^{2}$ facilitates further computations by allowing algebraic information pairs to be stored in formulaic forms. 
$\text{ISP}^{2}$ demonstrates good performance on mathematical datasets, enhancing the effectiveness of solutions by revealing key data essential for problem resolution, thus improving the performance of both ComplexCoT and Self Consistency.

Notably, as detailed in Table \ref{tab:performance_scores}, compared to the current SOTA plug-and-play methods in mathematics, our approach performs well, reaching new best levels in AddSub and SVAMP, and second-best levels in GSM8K and AQuA. 
The main reason is that $\text{ISP}^{2}$ continuously improves solutions by selectively gathering previous descriptions of problem space, allowing it to accurately resolve issues.
Instead of relying on multiple rounds of interactive reasoning, repetitive reading, or relationship extraction to obtain crucial information, solving many complex mathematical problems emphasizes the thought process and information filtering. 
By focusing on these aspects, misleading information is excluded, and effective problem-solving strategies are developed. 
It enables $\text{ISP}^{2}$ to achieve significant improvements in mathematical reasoning.

\paragraph{Commonsense reasoning} 

\begin{figure}
	\centering
	\includegraphics[width=3.5in]{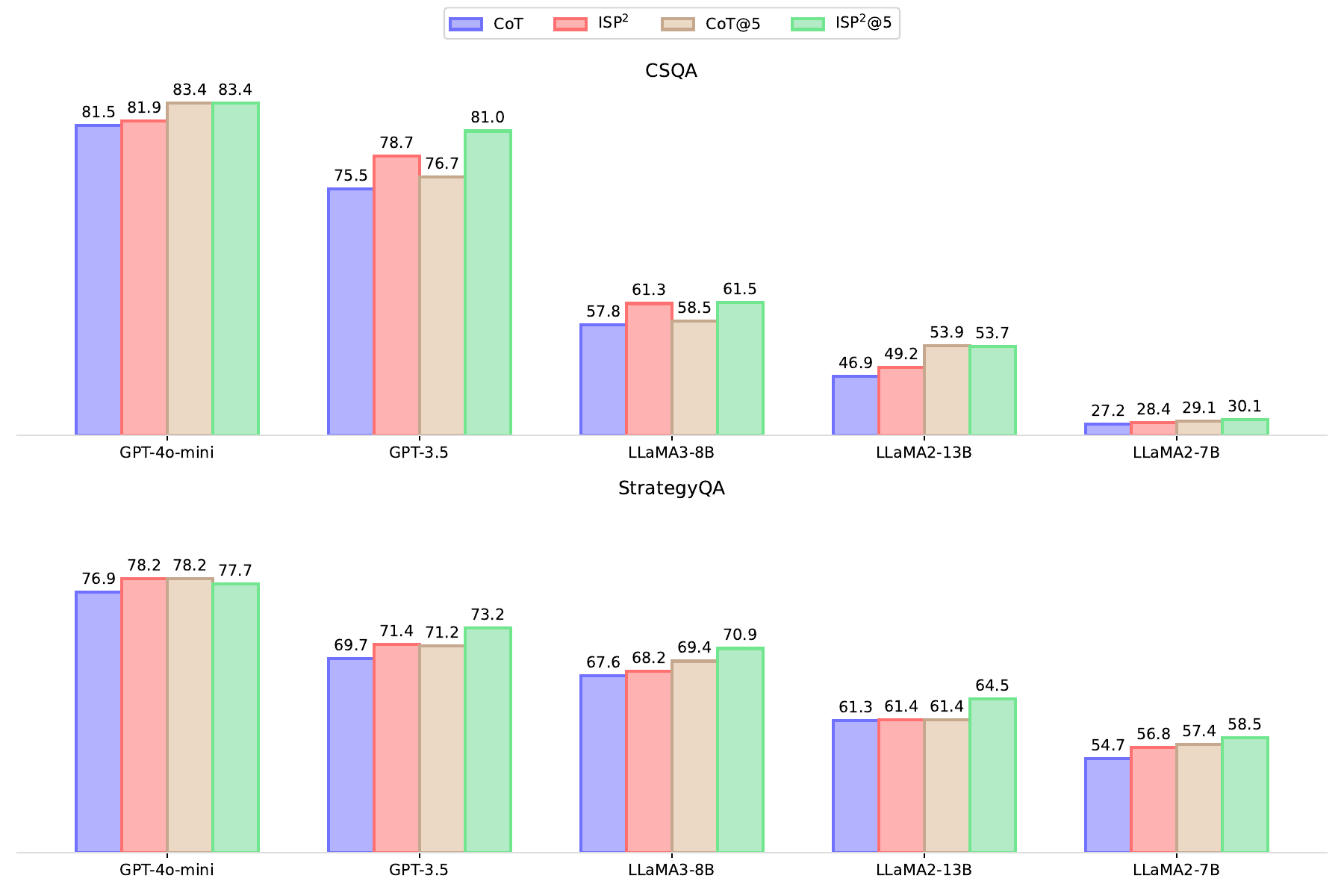}
	\caption{Result on Commonsense Problem.}
	\label{fig10}
\end{figure}

As shown in Figure  \ref{fig10}, for the StrategyQA and CommonsenseQA datasets, $\text{ISP}^{2}$ has increased performance by $4.9\%$ and $2.8\%$, respectively. 
Additionally, Self Consistency maintains an orthogonal relationship with $\text{ISP}^{2}$, both enhancing the performance of CoT, enabling it to better understand common sense and make more accurate choices in commonsense reasoning. 
There is a significant performance improvement in the CSQA dataset, where many hidden pieces of information exist within the problems, and relying solely on explicit knowledge is insufficient for accurate responses. 
Notably, in StrategyQA with the LLaMA2 model, $\text{ISP}^{2}$ did not demonstrate significant improvement. 
Smaller parameter LLMs have a tendency to include irrelevant text when forming information pairs. 
Consequently, this inclusion leads to the formation of continuous erroneous reasoning chains. 
Such errors can impede the ability of $\text{ISP}^{2}$ to effectively enhance performance.
However, most experiments still show that $\text{ISP}^{2}$ outperforms most base prompts in commonsense datasets, demonstrating strong capability of $\text{ISP}^{2}$ to guide CoT. 
When compared with existing SOTA methods, it performs comparably to ERA in SQA, which excels at controlling the reasoning process using entity relations, and also maintains second-best performance in CSQA.

\subsection{Ablation Studies}

$\text{ISP}^{2}$ involves multiple processes for handling information pairs and summarizing knowledge when assisting CoT predictions. 
We break down various steps to assess the impact of each component in $\text{ISP}^{2}$ on model performance, which helps us understand the importance of different factors within $\text{ISP}^{2}$. 
The various variants are listed below:
\begin{itemize}
	\item \textbf{Only Entity Extraction (OE):} This variant represents that before reasoning to answer questions, we only involve using LLMs for named entity recognition to extract entities as useful information. It omits the step of iterative summarization and directly provides the entities to LLMs for reasoning.
	\item \textbf{Only Information Pair (OIP):} This variant differs from OE in that we perform a complete extraction of information pairs during information retrieval, yet still omit the step of iterative summarization, directly providing the entities to LLMs for reasoning.
	\item \textbf{Score Alteration-$\text{ISP}^{2}$ ($\text{SAISP}^{2}$):} In this variant, we retain the $\text{ISP}^{2}$ steps, but in the iterative summarization, we change from summarizing the two information pairs with the lowest scores to those with the highest scores.
\end{itemize}

\begin{table}[!b]
	\centering
	\caption{Ablation study on $\text{ISP}^{2}$ aiding CoT reasoning.}
	\label{experiment_12}
	\resizebox{3.5in}{!}{
		\begin{tabular}{|c|c|cccccc|c|}
			\hline
			\multirow{2}{*}{Model} & \multirow{2}{*}{Method} & \multicolumn{6}{c|}{Dataset} & \multirow{2}{*}{Average}\\
			\cline{3-8}
			& & AddSub & SVAMP & GSM8K & AQuA & CSQA & SQA &\\
			\hline
			\multirow{4}{*}{GPT-3.5 Turbo}    & OE & 84.3 & 82.4 & 79.8 & \underline{60.2} & 75.1 & \underline{71.3} & 75.52\\
			&OIP & \underline{85.7} & \underline{86.3} & \underline{80.4} & 60.1 & \underline{76.7} & 70.1 & \underline{76.55}  \\
			&$\text{SAISP}^{2}$ & 83.8 & 85.1 & 76.9 & 59.9 & 75.9 & 69.1 & 75.12\\
			&$\text{ISP}^{2}$ & \textbf{88.6} & \textbf{88.7} & \textbf{83.4} & \textbf{64.4} & \textbf{78.7} & \textbf{71.4} & \textbf{79.20}  \\
			\hline
			\multirow{4}{*}{LLaMA2-13B}    & OE & 58.5 & 36.7 & 14.1 & \underline{18.9} & 45.2 & 60.3 & 38.95 \\
			&OIP & \textbf{61.2}  & \textbf{40.2} & \underline{17.2} & \underline{18.9} & \underline{46.6} & \textbf{61.7} & \underline{40.97} \\
			&$\text{SAISP}^{2}$ & 59.2 & 37.2 & 15.9 & \underline{18.9} & 43.9 & 60.3 & 39.23 \\
			&$\text{ISP}^{2}$ & \underline{60.0} & \underline{39.2} & \textbf{19.3} & \textbf{20.4} & \textbf{49.2} & \underline{61.4} & \textbf{41.58} \\
			\hline
			\multirow{4}{*}{LLaMA2-7B}    & OE & \underline{34.9} & 37.5 & \textbf{11.1} & 19.8 & \underline{27.7} & \textbf{57.2} & \underline{31.37}\\
			&OIP & 34.6 & \underline{38.5} & \underline{10.6} & \underline{20.1} & 27.5 & 56.5 & 31.30\\
			&$\text{SAISP}^{2}$ & 34.6 & 37.8 & 9.4 & 19.8 & \underline{27.7} & 56.1 & 30.91\\
			&$\text{ISP}^{2}$ & \textbf{35.2} & \textbf{39.1} & 8.3 & \textbf{21.6} & \textbf{28.4} & \underline{56.8} & \textbf{31.57}\\
			\hline
		\end{tabular}
	}
\end{table}

\begin{table}
	\centering
	\caption{Ablation study on $\text{ISP}^{2}$ aiding CoT reasoning using Self-Consistency.}
	\label{experiment_4}
	\resizebox{3.5in}{!}{
		\begin{tabular}{|c|c|cccccc|c|}
			\hline
			\multirow{2}{*}{Model} & \multirow{2}{*}{Method} & \multicolumn{6}{c|}{Dataset} & \multirow{2}{*}{Average}\\
			\cline{3-8}
			& & AddSub & SVAMP & GSM8K & AQuA & CSQA & SQA &\\
			\hline
			\multirow{4}{*}{GPT-3.5 Turbo}    & OE@5 & 90.1 & 85.8 & 83.2 & 66.8 & \underline{72.4} & 77.4 & 79.28\\
			&OIP@5 & 92.5  & 85.7 & 83.2 & \underline{69.9} & 72.3 & 78.2 &\underline{80.30}  \\
			&$\text{SAISP}^{2}$@5 & \underline{93.4} & \underline{87.1} & \underline{84.2} & 67.2 & 71.1 & \underline{78.6} & 80.27\\
			&$\text{ISP}^{2}$@5 & \textbf{93.9} & \textbf{90.1} & \textbf{84.8} & \textbf{72.7} & \textbf{73.2} & \textbf{81.0} &\textbf{82.62}  \\
			\hline
			\multirow{4}{*}{LLaMA2-13B}    & OE@5 & 62.0 & 48.1 & 18.7 & 22.3 & 52.3 & 63.4 & 44.47 \\
			&OIP@5 & \textbf{64.6} & \underline{48.2} & \textbf{22.1} & 22.9 & \underline{53.1} & \underline{63.7} & \underline{45.77} \\
			&$\text{SAISP}^{2}$+@5 & 61.0 & 48.0 & 20.9 & \underline{24.5} & 52.6 & 61.8 & 44.80 \\
			&$\text{ISP}^{2}$+@5 & \textbf{64.6} & \textbf{49.7} & \underline{21.9} & \textbf{25.4} & \textbf{53.7} & \textbf{64.5} & \textbf{46.63} \\
			\hline
			\multirow{4}{*}{LLaMA2-7B}    & OE@5 &\underline{42.0} & 29.3 & \underline{11.3} & \underline{26.3} & \underline{29.2} & 55.6 & 32.28\\
			&OIP@5 & 40.1 & \underline{7.8} & 8.9 & 24.0 & 28.9 & \underline{55.8} & \underline{32.58}\\
			&$\text{SAISP}^{2}$@5 & 38.5 & 36.1 & 8.1 & 23.2 & 27.4 & 54.6 & 31.32\\
			&$\text{ISP}^{2}$@5 & \textbf{43.9} & \textbf{42.5} & \textbf{11.4} & \textbf{29.9} & \textbf{30.1} & \textbf{58.5} & \textbf{36.05}\\
			\hline
		\end{tabular}
	}
\end{table}

Information pair extraction and two low-score information pair summarization are effective for answering questions.
In our experiments across six datasets, both CoT and Complex CoT methods showed performance improvements when entities and complete information pairs were incorporated. 
The inclusion of complete information pairs (OIP) provided significantly greater benefits for problem reasoning compared to entity extraction (OE). 
OIP provides a structured contextual description that enables the model to directly focus on the problem-solving logic without being distracted by additional context understanding. 
In contrast, OE merely extracts scattered keywords or entities, forcing the model to reconstruct fragmented information during reasoning.
OE not only increases cognitive load but also makes the model prone to errors due to misinterpretation of the context.

The most intriguing finding was the underperformance of ISP$^2$ when selecting the two highest-scoring information pairs for iterative summarization. 
This counter-intuitive result can be attributed to the inherent limitations of high-scoring pairs. 
While these pairs are prioritized for their surface-level relevance, their utility often saturates early during summarization. 
For instance, in mathematical reasoning tasks, high-scoring pairs might repeatedly emphasize a single solution path without addressing edge cases or alternative formulations. 
The early convergence leaves little room for deeper exploration, ultimately restricting the model’s ability to refine its reasoning.

In contrast, selecting low-scoring information pairs yielded superior results. 
These pairs, though initially deemed less relevant, often contain indirect or non-obvious associations. 
Their diversity introduces latent constraints or cross-domain connections that enrich the reasoning process. 
Iterative summarization of low-scoring pairs creates a self-correcting feedback loop: if one pair suggests an erroneous solution path, subsequent pairs may introduce conflicting evidence, prompting the model to reconcile discrepancies. 
The dynamic mimics human-like critical thinking, where hypotheses are refined through incremental validation. 
High-scoring pairs reduce uncertainty prematurely, increasing the risk of overcommitting to incorrect premises. 
Low-scoring pairs, by preserving diverse perspectives, act as a ``checks-and-balances'' system that balances exploitation of strong signals with exploration of implicit knowledge. 
The selection mechanism of low-score pair is particularly vital in complex tasks where superficial relevance must be weighed against deeper logical consistency.

\begin{table}
	\centering
	\caption{Ablation study on $\text{ISP}^{2}$ aiding Complex CoT reasoning.}
	\label{experiment_5}
	\resizebox{3.5in}{!}{
		\begin{tabular}{|c|c|cccc|c|c|}
			\hline
			\multirow{2}{*}{Model} & \multirow{2}{*}{Method} & \multicolumn{4}{c|}{Dataset} & \multirow{2}{*}{Average}\\
			\cline{3-6}
			& & AddSub & SVAMP & GSM8K &AQuA &\\
			\hline
			\multirow{4}{*}{GPT-3.5 Turbo}    & OE & \underline{86.6} & 83.9 & \underline{83.9} & 59.4 & 78.45 \\
			&OIP & 86.3 & \underline{85.3} & 82.8 & \underline{60.1} & \underline{78.63} \\
			&$\text{SAISP}^{2}$ & 85.2 & 84.9 & 80.1 & 58.7 & 77.23 \\
			&$\text{ISP}^{2}$ & \textbf{89.1} & \textbf{87.2} & \textbf{84.6} & \textbf{63.7} & \textbf{81.15} \\
			\hline
			\multirow{4}{*}{LLaMA2-13B}    & OE & 64.8 & \underline{44.8} & 15.7 & \textbf{20.4} & 36.43 \\
			&OIP & 66.8 & \underline{44.8} & \underline{17.7} & 19.2 & \underline{37.13} \\
			&$\text{SAISP}^{2}$ & \underline{69.1} & 43.1 & 16.5 & 17.7 & 36.60 \\
			&$\text{ISP}^{2}$ & \textbf{70.9} & \textbf{47.8} & \textbf{18.8} & \textbf{20.4} & \textbf{39.48} \\
			\hline
			\multirow{4}{*}{LLaMA2-7B} & OE & \underline{35.2} & 34.6 & 7.8 & 22.7 & 25.01 \\
			&OIP & \underline{35.2} & \underline{37.6} & \underline{8.4} & \underline{25.4} & \underline{26.65} \\
			&$\text{SAISP}^{2}$ & 33.4 & 35.0 & 7.4 & 23.8 & 23.89 \\
			&$\text{ISP}^{2}$ & \textbf{36.8} & \textbf{38.3} & \textbf{8.6} &\textbf{25.9} & \textbf{27.40} \\
			\hline
		\end{tabular}
	}
\end{table}

\begin{table}
	\centering
	\caption{Ablation study on $\text{ISP}^{2}$ aiding Complex CoT reasoning using Self-Consistency.}
	\label{experiment_6}
	\resizebox{3.5in}{!}{
		\begin{tabular}{|c|c|cccc|c|c|}
			\hline
			\multirow{2}{*}{Model} & \multirow{2}{*}{Method} & \multicolumn{4}{c|}{Dataset} & \multirow{2}{*}{Average}\\
			\cline{3-6}
			& & AddSub & SVAMP & GSM8K &AQuA &\\
			\hline
			\multirow{4}{*}{GPT-3.5 Turbo}    & OE@5 & 91.2 & 84.9 & 84.2 & 68.9 & 82.30 \\
			&OIP@5 & 92.2 & 86.7 & 85.2 & 68.8 & 83.23 \\
			&$\text{SAISP}^{2}$@5 & \underline{92.3} & \underline{88.6} & \underline{86.9} & \underline{69.2} & \underline{84.25} \\
			&$\text{ISP}^{2}$@5 & \textbf{92.8} & \textbf{90.8} & \textbf{87.0} & \textbf{70.5} & \textbf{85.28} \\
			\hline
			\multirow{4}{*}{LLaMA2-13B}    & OE@5 & 64.8 & 56.4 & 17.9 & 23.7 & 40.70 \\
			&OIP@5 & 66.8 & \textbf{57.1} & 19.8 & \textbf{24.8} & 42.13 \\
			&$\text{SAISP}^{2}$@5 & \underline{72.2} & 55.6 & \underline{20.2} & \underline{24.5} & \underline{43.13} \\
			&$\text{ISP}^{2}$@5 & \textbf{74.9} & \textbf{57.1} & \textbf{20.4} & 24.4 & \textbf{44.20} \\
			\hline
			\multirow{4}{*}{LLaMA2-7B}    & OE@5 & \textbf{43.8} & \underline{42.1} & \underline{11.4} & 20.1 & \underline{29.35} \\
			&OIP@5 & 40.3 & 34.9 & 9.7 & \underline{24.4} & 27.33 \\
			&$\text{SAISP}^{2}$@5 & 40.6 & 36.1 & 6.3 & \underline{24.4} & 26.85 \\
			&$\text{ISP}^{2}$@5 & \underline{43.5} & \textbf{43.6} & \textbf{12.8} & \textbf{27.2} & \textbf{31.78} \\
			\hline
		\end{tabular}
	}
\end{table}

\section{Discussion}

\subsection{Summarization Steps Analysis}
\begin{figure}
	\centering
	\includegraphics[width=3.5in]{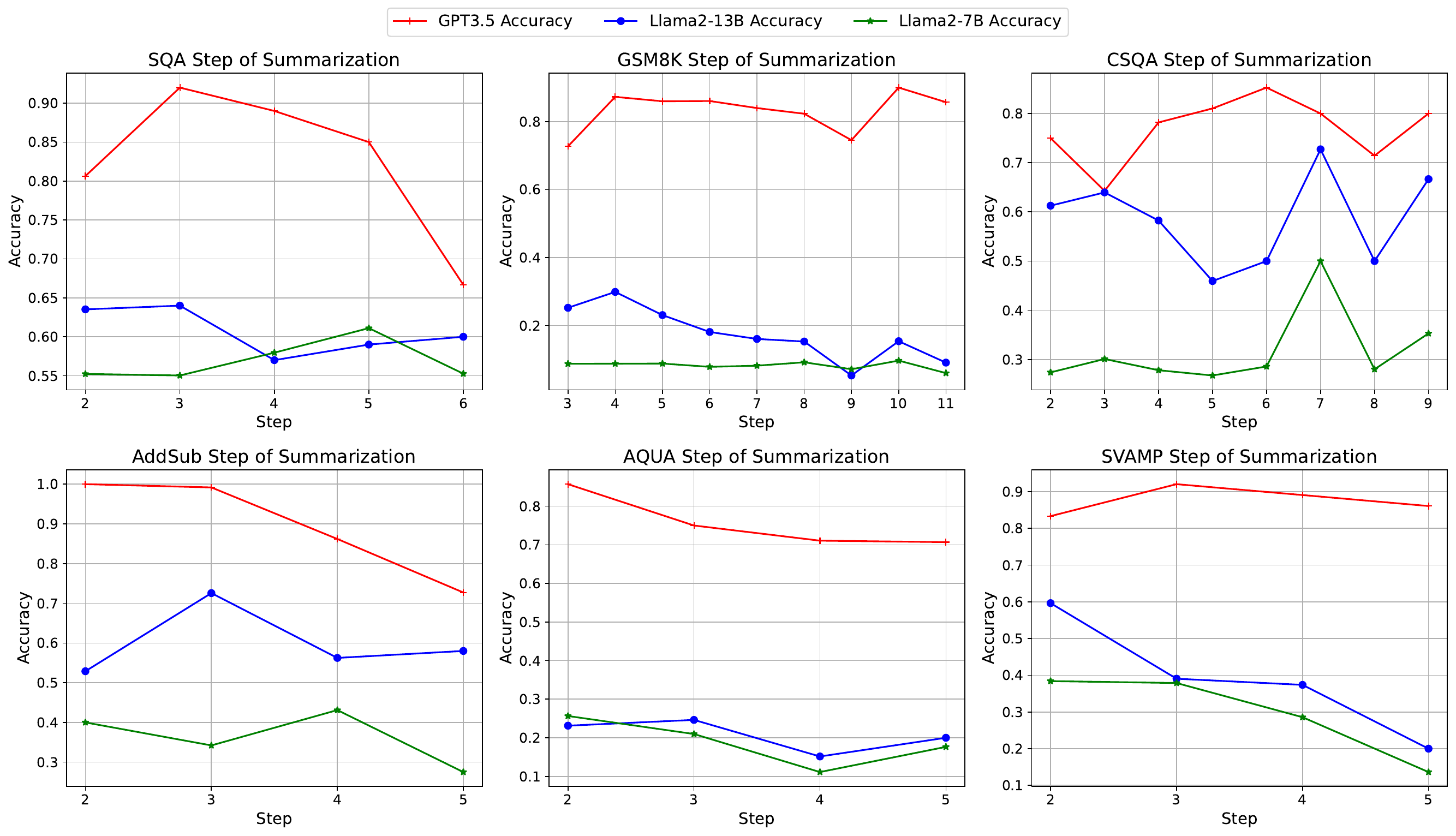} 
	\caption{Distribution of summarization steps and accuracy across different LLMs on various datasets}
	\label{fig4}
\end{figure}

We analyze the distribution of iterative summarization step lengths during inference and their positive guiding effect on reasoning.
In Figure \ref{fig4}, we illustrate the impact of final information pairs generated from different step lengths on the performance of LLMs across six datasets.
A common observation is that shorter steps consistently provide effective information for answering questions. 
We also find that the step length distribution for each task predominantly falls within the category of short steps. 
During adaptive extraction, much of the less helpful information has already been filtered out, resulting in step length compression and simplifying the reasoning process.
In fact, on LLaMA, information generated with longer steps can still effectively assist reasoning. 
However, for GPT-3.5, longer steps may not offer substantial support. 
We believe this is because GPT inherently possesses strong problem-solving capabilities, and excessively long steps might interfere with the coherence of final information integration.
In contrast, LLaMA can leverage more extensive summarization steps to enhance the processing and retention of knowledge within the information. 
These summaries serve as navigation tools within the problem space, continuously reinforcing and accumulating critical details to guide the transition from the initial state to the goal state.

\subsection{Error Source Analysis}

We extract 100 erroneous samples from each dataset and identified three critical types of errors that arise during the $\text{ISP}^{2}$ process. 
1. Information Pair Error (IPE): Failure to extract all explicit information pairs in the question or extraction of information pairs with misleading information;
2. Summarization Error (SE): Summaries that contain misleading information;
3. Reasoning Error (RE): Correct summarization but generation of an incorrect answer due to faulty reasoning.
These types of errors represent potential points of failure at each step of the $\text{ISP}^{2}$ process, which can propagate and affect subsequent operations.

\begin{table}[ht]
	\centering
	\caption{Proportion of different error categories across various datasets.}
	\label{experiment_3}
    \resizebox{3in}{!}{
    	\begin{tabular}{@{}p{2.5cm}p{2.2cm}ccc@{}}
    		\toprule
    		\textbf{Task}                & \textbf{Dataset}    & \textbf{IPE} & \textbf{SE} & \textbf{RE} \\ \midrule
    		Commonsense                  & StrategyQA               & 24\%         & 20\%         & 15\%         \\
    		Reasoning                    & CSQA                        & 19\%         & 22\%         & 10\%         \\ \midrule
    		& SVAMP                      & 19\%         & 11\%         & 3\%          \\
    		Mathematical                 & AQuA                     & 38\%         & 12\%         & 10\%         \\
    		Reasoning                    & AddSub                   & 10\%         & 17\%         & 13\%         \\
    		& GSM8K                        & 21\%         & 10\%         & 32\%         \\ \bottomrule
    	\end{tabular}
    }
\end{table}

Table \ref{experiment_3} displays the distribution of error categories across each dataset. 
Considering the error categories, the probability of information pair extraction errors is the highest, while the error rate for summarization is relatively lower, and the error rate for inference is the lowest.
We observe that in commonsense reasoning datasets, the error rates for information extraction and summarization are close. 
It is attributed to the fact that the model's own knowledge contains some elements related to implicit information, so the defects in information pair extraction are not significant.
However, during the final inference stage, conflicts often arise between the LLM's inherent knowledge and the already summarized information, leading to judgment errors in the final decision-making process.
The rate of information pair extraction errors in mathematical datasets shows a positive correlation with the complexity of the dataset. 
From basic arithmetic operations to complex algebraic calculations, this trend becomes increasingly pronounced. 
It indicates that as the difficulty of problems increases, the LLM's limitations in mathematical understanding become more apparent, leading to a greater impact on its ability to accurately collect and process mathematical information. 
Additionally, summarization helps improve dataset accuracy, but deficiencies in the large model's mathematical computation capabilities during the summarization process still exist.
Indeed, this issue can also arise during the inference process. 
Even if the summarized content is accurate, computational errors can still occur during inference.

\section{Conclusion}

We propose a new method called Iterative Summarization Pre-Prompting ($\text{ISP}^{2}$), which utilizes LLMs for precise information extraction and iterative summarization processes. 
By drawing on human approaches to understanding problems, $\text{ISP}^{2}$ enhances the reasoning capabilities of the CoT method when applied to LLMs.
The process of information understanding can address the weaknesses of these classical methods, providing a way to solve complex problems where the known information is too scattered, not cohesive, and not formalized. 
Additionally, $\text{ISP}^{2}$ can significantly improve the performance of LLMs on several datasets, and it can be easily combined with CoT and Self Consistency to further enhance reasoning effectiveness.
Equally important, the framework in our work only demonstrates the enhanced reasoning capabilities of $\text{ISP}^{2}$. 
From a broader perspective, we consider this an invitation to expand inquiry. 
We hope our method will inspire further research in NLP. 
It revisits the discussion of problem space interpretation, incorporating Simon's linguistic theories. 
$\text{ISP}^{2}$ should provide valuable references for researchers, encouraging them to undertake more in-depth investigations across a variety of languages.

\section{Acknowledgments}
\noindent This work was supported in part by the Science and Technology Commission of Shanghai Municipality under Grant (21DZ2203100), in part by the National Natural Science Foundation of China under Grant (62006150).

\bibliographystyle{IEEEtran}
\bibliography{references}

\end{document}